\pdfoutput=1

\documentclass[11pt,twocolumn]{article}

\usepackage{ifthen}

\newboolean{aclsubmission}
\setboolean{aclsubmission}{false}  

\ifthenelse{\boolean{aclsubmission}}{
    \usepackage{ACL2023}  
    \aclfinalcopy  
}{
    \usepackage{geometry}
    \geometry{margin=1in}
    \usepackage{authblk}
}

\usepackage{times}
\usepackage{latexsym}

\usepackage[T1]{fontenc}

\usepackage[utf8]{inputenc}

\usepackage{microtype}

\usepackage{inconsolata}

\usepackage{graphicx}
\usepackage{booktabs}
\usepackage{enumitem}
\usepackage{amsmath}
\usepackage{multirow}
\usepackage{subfig}

\title{Sentence-level Aggregation of Lexical Metrics Correlates \\ Stronger with Human Judgements than Corpus-level Aggregation}

\ifthenelse{\boolean{aclsubmission}}{
    
    \author{Paulo Cavalin \\
      IBM Research \\
      \texttt{pcavalin@br.ibm.com} \\
      \And
      Pedro H. Domingues \\
      PUC-Rio \\
      \texttt{phd.engmec@gmail.com} \\
      \And
      Claudio Pinhanez \\
      IBM Research \\
      \texttt{csantosp@br.ibm.com} \\
      }
}{
    \author[1]{Paulo Cavalin}
    \author[2]{Pedro H. Domingues}
    \author[3]{Claudio Pinhanez}
    \affil[1]{IBM Research, \texttt{pcavalin@br.ibm.com}}
    \affil[2]{PUC-Rio, \texttt{phd.engmec@gmail.com}}
    \affil[3]{IBM Research, \texttt{csantosp@br.ibm.com}}
}

\begin{document}

\maketitle

\begin{abstract}
In this paper we show that corpus-level aggregation hinders considerably the capability of lexical metrics to accurately evaluate machine translation (MT) systems. With empirical experiments we demonstrate that averaging individual segment-level scores can make metrics such as BLEU and chrF correlate much stronger with human judgements and make them behave considerably more similar to neural metrics such as COMET and BLEURT. We show that this difference exists because corpus- and segment-level aggregation differs considerably owing to the classical average of ratio versus ratio of averages Mathematical problem. Moreover, as we also show, such difference affects considerably the statistical robustness of corpus-level aggregation. Considering that neural metrics currently only cover a small set of sufficiently-resourced languages, the results in this paper can help make the evaluation of MT systems for low-resource languages more trustworthy.
\end{abstract}

\section{Introduction}
We can currently group machine translation (MT) metrics into two main groups: lexical and neural metrics \cite{freitag-etal-2023-results}. While lexical metrics such as the BLEU score \cite{papineni-etal-2002-bleu_} have considerably contributed to the progress in MT in the past 20 years, neural metrics emerged in the past few years as viable alternatives to overcome not only the shortcomings of lexical matches of n-grams, but also to leverage corpus-based training to improve MT evaluation \cite{rei-etal-2020-comet_,sellam2020bleurt,bert-score}.

Although it is quite clear that neural metrics are more robust and will eventually replace the lexical ones \cite{mathur-etal-2020-tangled,freitag-etal-2021-results,freitag-etal-2022-results}, in this paper we argue not only that lexical metrics are still quite needed but also that there is room to improve the robustness of such metrics. We say that they are needed because most progress with neural metrics is observed on the one hundred or so most-resourced languages in the world, but almost 7,000 languages in the world still lack the minimum amount of data to train a MT model \cite{lorandi-belz-2024-high}. It is thus unrealistic to think that a neural metric will be applied for such scenarios in the near future. With this perspective in mind, we argue that there is a very under-explored and important component of lexical metrics, which is the aggregation method.

Lexical metrics such as BLEU and chrF usually rely on corpus-level aggregation (CLA) \cite{papineni-etal-2002-bleu_,popovic-2015-chrf_}, but one can easily rely on segment-level aggregation (SLA). The main difference between CLA and SLA is that, while in the former we compute n-gram matching statistics for all samples in a first step and then we compute a global score for the entire test set, with the latter we compute the statistics and the score for each sample individually and then use the mean of these scores for evaluating a test set. We notice that SLA is a very under-explored method for lexical metrics (see Appendix~\ref{sec:review_mt_papers} for detailed statistics) and, for the best of our knowledge, there is no previous work demonstrating why corpus-level should be preferred besides theoretical assumptions.

At a glance, it surely looks that, in the worst-case-scenario, both CLA and SLA present comparable results, so there would be no good reason to question the aggregation method choice. But in this paper we show that, counter-intuitively against the common belief that CLA is better and should be the aggregation method of choice for lexical metrics, SLA is far more correlated to human judgements and to more robust neural metrics. To support this claim, we first show that there is a conceptual difference between these two aggregation methods, which we can mathematically relate to the \emph{average of ratios vs ratios of averages} problem. And based on empirical experiments, we show not only that the choice between the aggregation method significantly impacts the resulting system-level scores but also that CLA is not statistically robust.

In greater details, in a first set of experiments we investigated whether the aforementioned mathematical differences between CLA and SLA statistically impact the scores provided by the metrics. For this, we considered 492 system outputs from the WMT23 metrics shared task \cite{freitag-etal-2023-results}, and computed system-level scores considering not only these two aggregation methods with both BLEU and chrF as base metrics but also scores computing the mean of bootstrap-resampled scores (BRS), as a reference point of a more statistically-robust approach. For this evaluation, we computed the pairwise Pearson correlation of the scores of each pair of metric and the results clearly show not only that the scores from CLA and SLA differ considerably but also that the scores from SLA correlate quite stronger with those from BRS.

We then conducted a deeper investigation of the statistical robustness of the two aggregation methods, where we focused on evaluating the impact of the size of the test set on the correlation of scores. For that, we relied on downsampled test sets and computed the correlations of scores from downsampled versions of both CLA and SLA against each other, and against the statistically-robust BRS. The results corroborated our previous findings that SLA is not only more statistically robust than CLA but also show that this method is as statistically robust as BRS and can replace it as a much computationally-cheaper alternative.

However, the most surprising and relevant result, in our opinion, is that CLA is not only statistically weaker than SLA but it actually lacks any statistical robustness. In other words, when compared to BRS, the correlations of CLA scores computed on larger test sets are quite close to those computed on a test set with only a single sample. That means that the corpus-level evaluation could be simply replaced by single-sample evaluations.

Finally, in order to materialize what actually means the previous mathematical and statistical differences between the aggregation methods, in terms of the impact on the resulting quality of system-level scores, we computed correlations between the metrics and human judgements. For this, we considered human annotations from the WMT23 Metrics Shared Task, and we also included additional neural metrics, i.e. COMET, BLEURT, and BERTScore, to provide a better view on the impact of the aggregation method. Our results provide strong evidence that SLA correlates much stronger with human judgements, and are much more comparable to the outcomes of BERTScore. Considering that BERTScore is the only neural metric among these three which does not takes into account the input sentence to compute the score, which is compatible to the way lexical metrics work, we believe that our results show that the use of segment-level aggregation reduces considerably the gap between lexical and neural metrics.

\section{Related work}
The most well-known \emph{lexical metric} for machine translation evaluation is the BLEU score, introduced more than two decades ago as a solution to make the development of MT system more scalable \cite{papineni-etal-2002-bleu}. The idea was to take advantage of a set of translations created by humans and to somehow measure the discrepancy between the outputs generated by a MT system and the reference translations. With that approach, one could develop different systems and select the one which produced the highest BLEU score in a completely automated fashion.

The way BLEU works is based on computing a Precision-like metric on overlaps of \emph{n-grams} between the MT outputs and the references. That is done by counting up the number of n-grams generated in the MT outputs that also appear in the references. This computation is then heuristically refined to address some issues such as wrongly-generated repetitions and very short texts, and to combine different n-gram levels. BLEU also inspired other popular lexical metrics such as \emph{chrF}~\cite{popovic-2015-chrf} and \emph{chrF++}\cite{popovic-2017-chrf}, which play important roles to expand current NLP efforts into low-resource languages.

Despite the wide adoption of BLEU for about two decades, several works have focused on exploiting and overcoming its limitations \cite{graham-baldwin-2014-testing,mathur-etal-2020-tangled,freitag-etal-2022-results,freitag-etal-2023-results}. One issue already tackled by the community is the reliance in a large set of parameters and lack of standard and transparency in reporting results \cite{post-2018-call_}.
But another limitation, which is the reliance on lexical matches, resulted in the proposal of different alternative metrics, notably \emph{neural metrics} such as COMET \cite{rei-etal-2020-comet}, BLEURT \cite{sellam2020bleurt}, and UniTE \cite{wan-etal-2022-unite}. 

As outlined in the WMT{22} shared task results \cite{freitag-etal-2022-results}, across diverse domains and tasks, neural-based metrics like MetricX XXL \cite{juraska-etal-2023-metricx}, COMET-22 \cite{rei-etal-2022-searching}, UniTE \cite{wan-etal-2022-unite} and BLEURT-20 \cite{sellam-etal-2020-bleurt} consistently outperformed BLEU and other non-neural counterparts in capturing evaluation nuances.
In the subsequent WMT{23} shared task \cite{freitag-etal-2023-results}, the evaluation framework has been enhanced, expanding the metrics set and relying on a global score calculated through a weighted average across tasks. The results underscored the better alignment of neural-based metrics with human judgments than with non-neural ones.

Nevertheless, it is worth highlighting that neural metrics come with additional cost. Some metrics such as UniTE and COMET compute scores by relying also on the input provided to generate the translation, which obviously limits the application to cases where both the source and the target languages were used to train the underlying model. Even the neural metrics that consider only the MT outputs and the reference to compute the scores are quite limited, since they are usually trained with just dozens of languages. That limits neural metrics to, at best, hundreds of languages, the high- and mid-resourced ones.

Considering the currently-limited application of neural metrics and the vast number of under-studied languages in the world, there is still a vast application field for metrics based on overlaps of n-grams such as BLEU and chrF. At the same time, we observe here that there is a gap on a better understanding on the shortcomings of corpus-level aggregation, considering that most of the recent metrics rely on averages of segment-level scores. 

As we show in Appendix~\ref{sec:review_mt_papers}, from 345 papers we inspected, only one relied on segment-level aggregation for the BLEU metric \cite{chen-etal-2023-blaser}. And we can find the use of segment-level aggregation for chrF in very few works, such as in the reports for the WMT Metrics shared task \cite{freitag-etal-2021-results,freitag-etal-2022-results, freitag-etal-2023-results}. What seems to be missing is a work  showing the impact of the choice of the aggregation method. This paper aims to bridge this gap.

\section{Corpus- vs Segment-level Aggregation}
In this section we focus on describing in details both corpus-level aggregation (CLA) and segment-level aggregation (SLA) methods, and on explaing why the choice of one over another is mathemathically different and might affect the resulting scores. For the sake of simplicity, we will focus on the BLEU score, but in our empirical evaluations presented afterwards we demonstrate that our hyphoteses are not limited to BLEU and are at least also applicable to chrF.

\subsection{A case study with BLEU}
With a case study with BLEU, in this section we discuss why we expect differences in the results according to the the aggregation method. For that we will rely on a simplified abstraction of BLEU, considering that this metric consists of computing a Precision-like score for the generated translation or for a set of generated translations \cite{papineni-etal-2002-bleu}.

Before explaining the aggregation method, it is worth explaining how BLEU is computed for a single sentence, i.e. the so-called \emph{sentence-level} BLEU score. For this case, we basically compute the total of \emph{n-gram matches} between the \emph{candidate} sentence, i.e. the sentence generated by the ML model, and the \emph{references}, representing the ground-truth translations generated by a person that is fluent enough in the target language. The matching is computed by evaluating the number of n-grams present in the candidate sentence that also appear in the references. After we computed that number of matches, we divide this number by the total of n-grams contained in the candidate output, to compute a Precision-ish score that represents
the sentence-level evaluation score. Of course, since this is a simplified abstraction, we are not taking into account modified n-gram precision, clipping, combination of different n-gram levels, and brevity penalty, but this does not affect our rationale.

When we expand sentence-level BLEU to evaluate the MT system on a corpus of text or on an entire test set, for instance, the main approach is to rely on the so-called \emph{corpus-level} BLEU (check Section~2.2.1 in \cite{papineni-etal-2002-bleu} for further details). The corpus-level aggregation (CLA) of BLEU consists of a global computation of n-gram matches and a single scoring for all samples in the test set, done at once. That is, all n-gram matches are counted and summed up, and this number is divided by the sum of the lengths of all candidate sentences.

Although corpus-level BLEU is the usual choice and the default options in tools such as SacreBLEU\footnote{https://github.com/mjpost/sacrebleu}, it is quite easy to adopt averages of segment-level scores, or simply \emph{segment-level aggregation} (SLA), to compute system-level scores \cite{bugliarello-etal-2020-easier,niu-etal-2020-evaluating}. The implementation is straightforward, where one simply need to compute sentence-level BLEU scores for each individual test sample, than using the mean of such scores as the final system-level scores. This aggregation method presents some advantages, such as allowing to compute statistical metrics such as standard deviations, which is not possible with corpus-level aggregation. Notice that bootstrap resampling is a somewhat popular method to compute statistical significance tests on corpus-level scores \cite{koehn-2004-statistical,jon-bojar-2023-breeding,fucci-etal-2023-integrating} and can also be used to compute statistical metrics, but it is a more expensive approach in terms of computation requirements.

At a first sight, it is reasonable to believe that CLA and SLA provide the same results, so the main advantage of the latter would only be the less expensive way to compute statistical metrics. But there is a conceptual mathematical difference between these two aggregation methods, which we explain next.



\subsection{CLA and SLA as differently weighted ratio averages}
\label{sec:bleu_vs_m-bleu_math}
We dive now into a mathematical explanation of the difference between corpus-level and segment-level aggregations, to demonstrate why these two methods may present differences in the results. We argue that the main difference between the two aggregation methods can be seen as a classical case of \textit{ratio of averages vs average of ratios}.

To understand this, let us adopt a simplified definition of BLEU as a ratio of the number of matches $m$ by n-grams $w$ in a corpus, as in the previous section. And let us also refer to the corpus-level and segment-level BLEU as simply BLEU and m-BLEU, respectively. Considering all the $n$ sentences $i$ of the corpus, it is evident that BLEU can be computed by the ratio between the sum of all partial matches $m_i$ in each sentence $i$ by the sum of n-grams in all sentences $i$, $w_i$:
\begin{equation}
    \mbox{BLEU} = \frac{m}{w} = \frac{\sum_{i=1}^{n}m_i}{\sum_{i=1}^{n}w_i} 
\end{equation}

Accordingly, m-BLEU is the average of ratios between the number of matches $m_i$ and the number of words $w_i$ of each sentence $i$:
\begin{equation}
    \mbox{m-BLEU} = \frac{1}{n} \sum_{i=1}^{n} \frac{m_i}{w_i} =  \sum_{i=1}^{n} \left(\frac{1}{n}\right) \frac{m_i}{w_i} 
\end{equation}

It is easy to derive that BLEU is the weighted average of the sentence ratios by the proportional length of each sentence~$i$: 
\begin{equation}
   \begin{aligned}
    \mbox{BLEU} & = \frac{\sum_{i=1}^{n}m_i}{\sum_{i=1}^{n}w_i}  = 
    \sum_{i=1}^{n} \frac{m_i}{w} 
     = \sum\limits_{i=1}^{n}  \frac{w_i}{w_i} \frac{m_i}{w} \\
 & \mbox{BLEU}    = \sum\limits_{i=1}^{n} 
        \left(\frac{w_i}{w}\right) \frac{m_i}{w_i}
     \end{aligned}
\end{equation}


As we see, m-BLEU weights the ratios equally with $1/n$ weights while BLEU weights the ratios with the value $w_i/w$ which is proportional to the length of each sentence~$i$. Therefore, BLEU results are likely to be biased by the proportion of matches and candidate sentences lengths, while m-BLEU considers the performance independently of that.

\section{Empirical evaluation}
\label{sec:empirical_evaluation}
In this section we present experiments aiming at investigating whether the choice of aggregation method actually impacts the score provided by a lexical metric. For that, we consider three different implementations of BLEU and chrF and, considering the outputs from 492 different systems, we present a detailed analysis on the distribution of scores provided by these metrics.

\subsection{The dataset}
For this investigation we rely on the WMT 2023 Metrics Shared Task dataset\footnote{https://wmt-metrics-task.github.io/}, or simply \emph{WMT23 dataset}, comprising the results of 492 different MT systems, involving different languages and domains. This dataset contains 468,850 system outputs with the corresponding inputs and references, in a quite diverse setting, containing 147 different language pairs, being 48 different source languages and 44 target languages. 

We converted the 468,850 raw entries to 492 system evaluations by grouping the data by dataset type (challengesets2023 or generaltest2023), dataset (challenge\_ACES, challenge\_DFKI, challenge\_NRC-MSLC23, or generaltest2023), language pair (147 options), and system (two systems for the challegeset2023 dataset type and 14 systems for the generaltest2023). Those are all inner structures of the dataset and the aggregated data will be made publicly available\footnote{anonimous link to dataframe}.

\subsection{Three implementations of BLEU}
\label{sec:different_bleus}
We considered three different implementations for BLEU. Two of them are based on the corpus- (CLA) and segment-level aggregation (SLA), and the third relies on bootstrap-resampled scores (BRS) for providing more robust statistical estimates, so that we can use this approach as a reference point for statistically-reliable scores.

As a consequence, the first implementation is referred to as simply {\bf BLEU}, consisting of the traditional BLEU score with CLA, here computed with the SacreBLEU tool with default parameters \cite{post-2018-call}\footnote{nrefs:1|case:mixed|eff:no|tok:13a|smooth:exp|version:2.0.0}. For this metric, given a test set with samples and the corresponding reference MT outputs, we take all MT outputs and references at once, in a single list, and compute the score with the \verb|corpus_bleu| function in Python code.

The second metric, referred to as {\bf m-BLEU}, implements SLA. For this we simply compute the score of each segment (i.e. a test sample) with SacreBLEU's \verb|sentence_bleu| function, again with default parameters, and calculate the overall average to provide the system-level score. 

The third metric, named {\bf x-BLEU}, consists of implementing the BRS method, as in \cite{niu-etal-2020-evaluating,liu-etal-2021-scheduled-sampling}. This is an alternative to computing system-level scores with higher statistical robustness, where the resamplings represent varied rearrangements of the test set to compute corpus-level BLEU. We rely on 1,000 resamplings, with replacement, of 1,000 samples\footnote{Notice that 1,000 samples is roughly very close the mean of samples in the WMT23 dataset, which is 952.94.} for each system, and we applied the same SacreBLEU's \verb|corpus_bleu| function, with default parameters, on top of each resampled set, generating 1,000 scores for each system. We then provide the average of those 1,000 scores as the system-level score.

\subsection{Three implementations of chrF}
In order to investigate whether our observation also generalize to other lexical metrics, we explored also three different implementations of chrF. Notice that chrF is quite similar to BLEU, where overlaps of characters n-grams are computed instead of overlap of words, and the final score is based on an F1 score instead of the precision-ish score used by BLEU. More importantly, the same variety of aggregation methods are possible to be used with chrF, and those varieties can be put in practice with the \verb|corpus_chrf| and a \verb|sentence_chrf| functions from SacreBLEU. 

Consequently, the three different chrF implemenations that we consider in this work are: {\bf chrF}, the CLA version computed with \verb|corpus_chrf| SacreBLEU's function; {\bf m-chrF}, the SLA implementation relying on the average of segment-level scores computed with \verb|sentence_chrf|; and {\bf x-chrF}, the BRS metric computed with averages of \verb|corpus_chrf| on 1,000 resamplings with 1,000 samples. Notice that we always rely on the default SacreBLEU's parameters for such functions.

\subsection{CLA and SLA correlate weakly to each other, and SLA correlates strongly to BRS}
\label{sec:bleu_only_evaluation}
Our first evaluation focused on investigating the distribution of scores provided by the three different aggregation methods, how these distributions correlate to each other, and how they correlate to more statistically-robust scores. We focus first on analysing BLEU, and then show results with chrF to understand the impact of the base metric.

\begin{figure}[t!]
    \centering
    \includegraphics[trim={0.5cm 0cm 0cm 0cm},width=5.5cm]{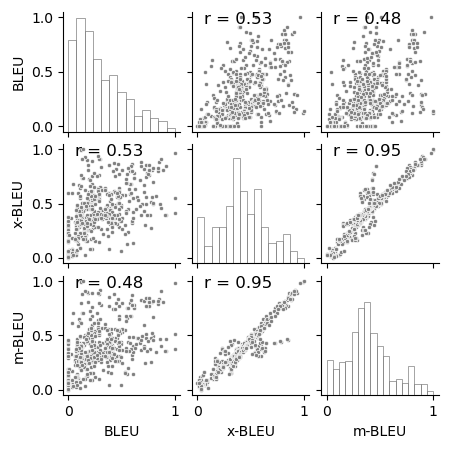}
    \caption{Correlation plots of the different BLEU metrics to each other and, in the diagonal, the distribution of the scores of the 3~metrics.}
    \label{fig:bleus_only}
\end{figure}

\begin{figure}[t!]
    \centering
    \includegraphics[trim={0.5cm 0cm 0cm 0cm},width=5.5cm]{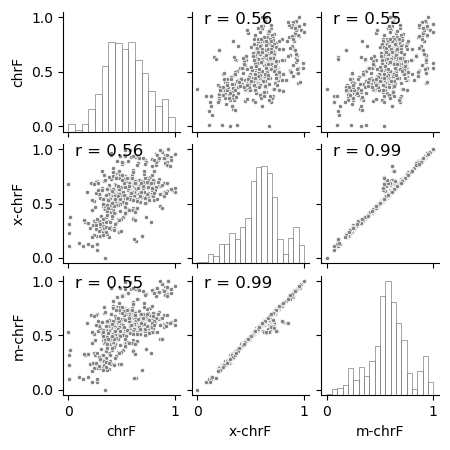}
    \caption{Correlation plots of the different chrF metrics to each other and, in the diagonal, the distribution of the scores of the 3~metrics.}
    \label{fig:chrfs_only}
\end{figure}
 
Figure~\ref{fig:bleus_only}, displaying a grid of 9~plots, presents the main results of this analysis. Notice that in the diagonal we show the histogram of the distribution of scores for each metric, considering the scores for the 492 systems, and in the upper and lower non-diagonal cells we present the pairwise scatter plots and Pearson correlations, in the -1 to 1 range, considering the metrics scores. The main intuition of computing Pearson correlation is to assess the linear correlation between the metrics. That is, a highly-positive correlation, i.e. a value close to 1.0, indicates that high scores in one metric correspond to high scores in another metric, and vice-versa. This correlation helps determine whether the metrics are aligned in capturing what are the good and bad translation results  and what is in-between.

From both the distributions and correlation values, we can clearly see that x-BLEU and m-BLEU tend to produce much closer results to one-another than to BLEU. We can see that the distribution of BLEU is right-skewed, while the other distributions are more centralized. In details, the means of the distributions are 0.29, 0.43, and 0.39, for BLEU, x-BLEU, and m-BLEU; and the correlations of BLEU to x-BLEU and m-BLEU are of 0.53 and 0.48, respectively, and x-BLEU and m-BLEU present 0.95 of Pearson correlation.

The results of a repetition of the experiments with chrF as the base metric is presented in Figure~\ref{fig:chrfs_only}, and we observe quite similar outcomes. That is, chrF correlates weakly with the other metrics and m-chrF correlates strongly to x-chrF. However, we observe that, unlike the BLEU metrics, the distributions of chrF, x-chrF, and m-chrF can be seen as centralized, presenting means of 0,54, 0.57, and 0.58, respectively. In terms of Pearson correlation, x-chrF and m-chrF present almost perfect correlations to each other with 0.99, while chrF correlates weakly to the other methods, with correlations of 0.56 and 0.55.

What it is very noticeable from these experiments, is that SLA presents a quite higher correlation to BRS than CLA does, which is strong evidence that SLA is not only as capable as BRS to compute statistically-robust scores but also a more-cost-effective option. Moreover, it also indicates that CLA is a quite less statistically-robust method.

\subsection{Aggregation methods present strong cross-metric correlation}
\begin{figure}[t!]
    \centering
    \includegraphics[trim={1cm 0cm 0cm 0cm},width=\columnwidth]{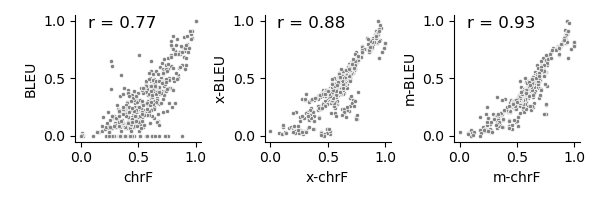}
    \caption{Correlation plots of the differeent implementations based on the BLEU and chrF metrics.}
    \label{fig:bleu-chrf}
\end{figure}

We now focus on investigating the correlations between the different aggregation methods of the different BLEU and chrF implementations. The experiments are analogous to those presented in the previous sections, so in Figure~\ref{fig:bleu-chrf} we present the scatter plot and the Pearson correlation of each pair of similar implementations, i.e. BLEU vs chrF, x-BLEU vs x-chrF, and m-BLEU vs m-chrF. 

Interestingly, these results indicate that similar aggregation methods correlate strong with each other. Although the correlation of BLEU to chrF is the weakest, with 0.77, it is above the 0.50 to 0.55 correlation values that such metrics presented to the non corpus-level ones. And the non corpus-level metrics correlate quite strong to each other, since x-BLEU vs x-chrF present a correlation of 0.88, and m-BLEU vs m-chrF of 0.93. These results indicate that, as we mentioned in Section~\ref{sec:bleu_vs_m-bleu_math}, corpus-level aggregation might biased by the ratio between n-gram matches and sentence lengths, and that can explain why BLEU and chrF correlate strong to one another, and weakly to the other methods.

\begin{figure}[t!]
    \centering
    \includegraphics[trim={0.5cm 0cm 0cm 0cm},width=\columnwidth]{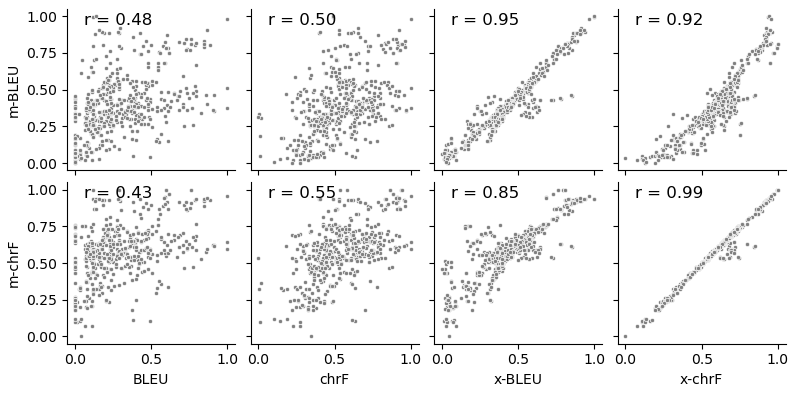}
    \caption{Correlation plots of the m-BLEU and m-chrF to the BLEU, chrF, x-BLEU, and x-chrF metrics.}
    \label{fig:m-to-others}
\end{figure}


In order to gather additional evidence, in Figure~\ref{fig:m-to-others} we present a cross-metric comparison betweem m-BLEU and m-chrF, against BLEU, chrF, x-BLEU, and x-chrF, respectively. Again, a weak correlation of BLEU and chrF to the non corpus-level metrics is seen, given that BLEU correlates weakly to x-chrF and m-chrF, and vice-versa. Moreover, m-BLEU and x-BLEU also correlate strongly to x-chrF and m-chrF, respectively, indicating that the corpus-level aggregation introduces a bias that can hinder the statistical robustness of lexical metrics.

\subsection{Characterizing the statistical robustness of CLA and SLA}
\label{sec:bleu_vs_downsampled_mbleu}
Given that one main outcome from the previous section is that corpus-level aggregation (CLA) metrics correlate weakly to their more statistically robust counterparts such as bootstrap-resampled scores (BRS), and that the segment-level aggregated (SLA) metrics correlate strongly to BRS, in this section we focus at investigating the statistical robustness of the aggregation methods. 

Our methodology consists of using BRS as an upper bound for statistical robustness for system-level scores, since they rely on the well-known bootstrap resampling method, and on not only evaluating the correlation between the distributions of BRS-based scores against downsampled versions of both CLA- and SLA-based scores, but also the correlation of these downsampled metrics to each other. With this approach we believe we can understand the sensitiveness of the aggregation methods to the number of samples and how their statistical robustness is affected by the number of samples.

\begin{figure}[t!]
    \centering
    \includegraphics[trim={0.5cm 0cm 0cm 0cm},width=\columnwidth]{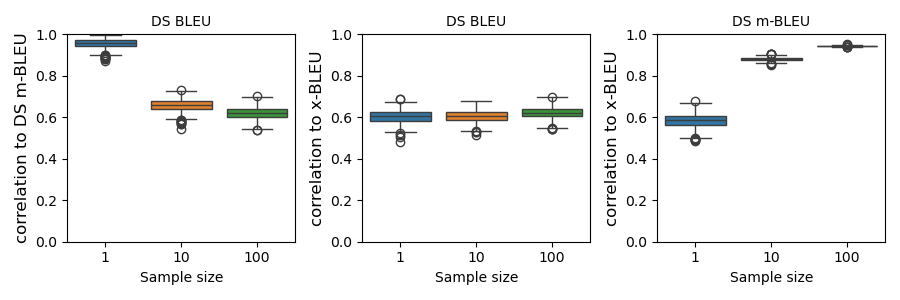}
    \includegraphics[trim={0.5cm 0cm 0cm 0cm},width=\columnwidth]{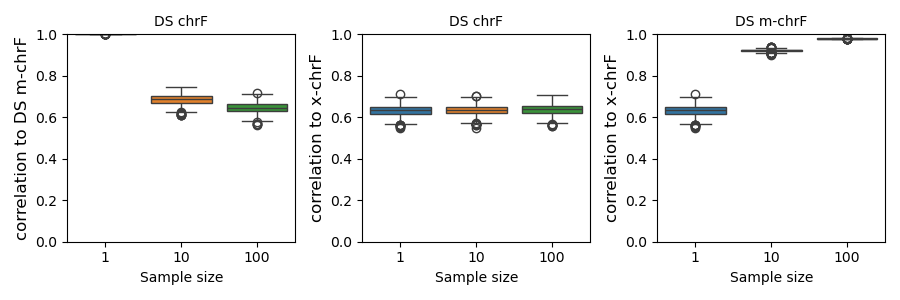}
    \caption{Boxplots of the correlations of downsampled (DS) m-BLEU scores to BLEU and x-BLEU, and of m-chrF scores to chrF and x-chrF.}
    \label{fig:downsampled_m-metrics}
\end{figure}

In greater detail, we created downsampled test sets for each of 492 datasets, considering three different sizes, i.e. $N=\{1, 10, 100\}$, and computed scores with both CLA- and SLA-based metrics on each of these downsampled test sets. Then, we computed the Pearson correlation between the scores of these downsampled version of each aggregation method and between these scores and the more statistically-robust ones computed with BRS. This process were repeated 1,000 times for a better statistical estimate. The distribution of correlations, for both BLEU- and chrF-related evaluations, are presented in Figure~\ref{fig:downsampled_m-metrics}.

The results show that CLA-based metrics are statistically weak. That is, as we can see in both top and bottom left-most plots, the downsampled CLA- and SLA-based metrics correlate strongly only in the case of datasets with only one sample, i.e. $N=1$, with correlation values approaching 1.0. Nevertheless, as the number of samples increases, their correlation reduces quite drastically, to about 0.7 with $N=10$ and about 0.6 with $N=100$. That means that, statistically speaking, we can claim that those metrics differ considerably to each other. 

We can go further and dare to claim that an evaluation of an entire test set with CLA is similar to conducting the evaluation of the same system with just a single sample. In the central plots of Figure~\ref{fig:downsampled_m-metrics}, we can see that the correlation of their downsampled scores to the more statistically-robust BRS-based methods does not increase, or increases very insignificantly, with the number of samples. On the other hand, as we can observe in the right-most plots of the same figure, the correlation of donwsampled SLA-based metrics correlate weakly with the bootstrapped scores with $N=1$, but the correlation increases significantly with $N=10$ and $N=100$, showing that this aggregation method does not suffer from the same lack-of-statistical-robustness problem.


\section{Impact of the aggregation method}
In order to materialize the actual impact of the aggregation method in computing system-level scores, in this section we present an evaluation comparing the scores of the metrics described in Section~\ref{sec:empirical_evaluation} compared to ground-truth data from human judgements. For that we rely on three language pairs with Multidimensional Quality Measurements (MQM), used to benchmark metrics for the WMT23 Metrics Shared Task \cite{freitag-etal-2023-results}. The human scores are computed with the weighted average of the multiple dimensions. To present a more extensive evaluation, we also consider the eight language pairs annotated with Direct Assessment (DA) scores\footnote{We relied on the DA assessments available in the wmt23 folder of the data downloaded with the code in https://github.com/google-research/mt-metrics-eval}.

By evaluating MQM and DA data individually, we compute the mean Pearson correlation of the system scores from each language pair to the implementation of BLEU and chrF described in Section~\ref{sec:empirical_evaluation}. Again, we consider plain Pearson correlation values, ranging from -1 to 1. For comparison purposes, we also include three neural metrics to present a better reference point related the impact of the aggregation method compared with these more robust metrics: COMET\footnote{Unbabel/wmt22-cometkiwi-da base model}
\footnote{https://github.com/Unbabel/COMET}, BLEURT\footnote{BLEURT-20 base model}\footnote{https://github.com/google-research/bleurt}, and BERTScore\footnote{bert-base-multilingual-cased base model}\footnote{https://github.com/Tiiiger/bert\_score}.

The results are presented in Table~\ref{tab:correlations_to_human_scores}, where we list the rankings of the metrics according to their correlation to the human-annotated data. The results, interestingly, show that the SLA-based methods, i.e. m-BLEU and m-chrF, not only considerably outperform BLEU and chrF, the CLA-based ones, but also that they provide correlations that are much closer to those of the neural metrics. We can also see that, with chrF and MQM annotations, we can improve from a moderate-to-low correlation of 0.392 to a strong correlation of 0.806, which is just 0.049 correlation points weaker than the correlation of BERTScore and just 0.066 of BLEURT. With the DA annotations, we can observe that m-chrF performs even closer to the neural metrics. Notice that the SLA-based metrics also outperform their BRS counterparts, i.e. x-BLEU and x-chrF, demonstrating again the statistical robustness of the SLA method.

\begin{table}[t!]
    \centering
    \caption{Rankings of Person correlations from metric scores to human judgements}
    \begin{small}
    \begin{tabular}{c||l|c||l|c||}
      & \multicolumn{2}{c||}{MQM} & \multicolumn{2}{c||}{DA} \\
      & metric & corr. & metric & corr. \\
      \hline
    1 & COMET & 0.923 & COMET & 0.926 \\
    2 & BLEURT & 0.872 & BLEURT & 0.917 \\
    3 & BERTScore & 0.855 & BERTScore & 0.821 \\
    4 & m-chrF & 0.806 & m-chrF & 0.802 \\
    5 & x-chrF & 0.804 & x-chrF & 0.793 \\
    6 & m-BLEU & 0.776 & m-BLEU & 0.729 \\
    7 & x-BLEU & 0.762 & x-BLEU & 0.456 \\
    8 & BLEU & 0.425 & chrF & 0.285 \\
    9 & chrF & 0.392 & BLEU & -0.006 \\
    \end{tabular}
    \end{small}
    \label{tab:correlations_to_human_scores}
\end{table}

\section{Final discussion}
In this work we presented a comparison of the traditional corpus-level aggregation against the less popular method based on averaging individual segment-level scores, showing that the latter can generate system evaluation scores which correlate stronger to human judgements and to neural metrics. We demonstrate that this difference happens because of a fundamental mathematical difference: CLA metrics are biased towards the performance on long sentences, considerably reducing the capability of lexical metrics to correlate with human judgements when corpus-level aggregation is considered. The main issue  we observed is that corpus-level aggregation voids the statistical robustness of a test set-based evaluation, providing scores that are comparable to evaluations with a single sample. 

One main outcome of this paper should be regarded as a clear recommendation of MT researchers: {\bf stop using corpus-level aggregation}. As we show, segment-level aggregation is not only better than corpus-level in terms of correlation to human judgments, but also comparable, if not better, than robust statistical evaluations based on bootstrap resampling. Similarly, MT researchers should {\bf use segment-level scores for statistical evaluation} instead of the expensive computations for bootstrap resampling, although one could also bootstrap resample segment-level scores to get even more robust estimates.

Finally, we would like to draw the attention for the vast application field that lexical metrics still have since neural metrics cover only about a hundred or so languages. Moreover, it is important to understand that some of the bad reputation which lexical metrics received in the past years might be under-deserved because of the wide, but wrong, adoption of corpus-level aggregation.

\section{Limitations}
One clear limitation of this work is relying on a single data source, which is the WMT23 Shared Task dataset. Nevertheless, since it is a very recent dataset, considerably large, and coming from a very well-known workshop on machine translation, we believe that this dataset is strong enough to experimentally provide evidence as used in our empirical evaluations.

Another limitation is to rely on a single tool to compute BLEU and chrF, which is SacreBLEU, even though there are other implementations available. Again, since this tool can be viewed as a de facto standard BLEU implementation~\cite{post-2018-call}, we believe that the tool is strong enough to experimentally prove our assumptions in the empirical evaluation.

Additionally, we have not thoroughly evaluated the metrics, in the sense of changing parameters of the metrics such as the maximum n-gram lengths and thus forth. We stayed with the default SacreBLEU's parameters only. But again, given the level of impact the changing the aggregation method presents, as we show, we do not believe that simply changing the base metrics' parameters would affect significantly the outcomes of this paper.

\section{Ethical considerations}
We are not aware of any ethical issue that this paper might raise. All of the data used for this research are publicly-available, and the outcomes of this paper are likely to contribute to improving the quality of the research of the whole machine translation community.

\ifthenelse{\boolean{aclsubmission}}{
    \bibliographystyle{acl_natbib}
}{
    \bibliographystyle{plain}
}
\bibliography{anthology,custom}

\clearpage
\appendix

\section{An comprehensive evaluation of of the use of lexical metrics in recent machine translation papers}
\label{sec:review_mt_papers}

\begin{table}[h!]
    \centering
    \caption{Metrics used in 345 recent MT papers (a single paper may use more than one metrics).}
    \label{tab:metrics_distribution}
    \begin{tabular}{lrr}
    \toprule
       Metric &  Total & \% \\
    \midrule
        BLEU  &   341 & 98.8 \\
        COMET &     36 & 10.4 \\
         chrF &     30 & 8.4 \\
       METEOR  &     17 & 4.9 \\
    BERTScore  &     15 & 4.3 \\
          TER &      15 & 4.3  \\
       BLEURT  &     10 & 2.9 \\
       BLONDE &      3 & 0.8\\
        UniTE &      3 & 0.8 \\
        Others &     17 & 4.9 \\
    \bottomrule
    \end{tabular}
\end{table}

\begin{table}[h!]
    \centering
    \caption{Distribution of BLEU implementations in 341 recent MT papers which used BLEU as a metric.}
    \label{tab:implementations_distribution}
    \begin{tabular}{lrr}
    \toprule
       Metric &  Total & \% \\
    \midrule
    BLEU (inferred)	& 210	& 60.1 \\
    Unclear	& 122	&  36.2 \\
    BLEU (explictly)	& 6	& 1.7 \\
    x-BLEU	& 3	& 0.9 \\
    m-BLEU	& 1	& 0.3 \\
    \bottomrule
    \end{tabular}
\end{table}

In this section we present an evaluation of the usage of lexical metrics in MT papers. To demonstrate the impact of our results, we analysed MT-related papers published in recent editions of \emph{ACL} and \emph{EMNLP}, two of the major venues for MT research, and catalogued the ways the evaluation of MT systems are reported. 

We inspected 345 papers which contained evaluations of MT systems published in the past four editions of ACL and EMNLP conferences, i.e. from 2020 to 2023. We selected those papers both by searching for terms such as \emph{Translation}, \emph{Machine Translation}, \emph{MT}, and \emph{BLEU} in the title, and by a further fine-grained manual inspection. We considered all papers that conducted evaluations of systems for MT-related tasks, an average of 43.12 per conference edition.

The distribution of metrics that appear in those papers is presented in Table~\ref{tab:metrics_distribution}. All but only 4 out of 345 papers did not report the use of BLEU, meaning that BLEU was employed by 98.8\% of papers we inspected. Also, the adoption of neural metrics is still low, with COMET being the metric with the most prominent level of usage, but only in 10.4\% of the papers. Notice that all neural metrics combined are reported in less than 1/4 of the papers, i.e. about 80 papers, considering also all the 17~metrics with less than 3 mentions grouped in the Others row.

Given the domination of the BLEU score in these papers, we then conducted a study on understanding what aggregation methods for this specific lexical metric are used and report the results in Table~\ref{tab:implementations_distribution}. Only 10 papers explicitly mention the aggregation method used for BLEU, where 6 of them explicitly mention that they are using corpus-level aggregation (simply BLEU in the table), 3 stated they are using averages of bootstrapped resamplings (x-BLEU), and only a single paper relies on averages of segment-level BLEU scores (m-BLEU). For the remaining 335 papers, we inferred the aggregation method by looking for references of specific tools they used, such as SacreBLEU \cite{post-2018-call} and Moses. We observed that 210 papers, 60.1\%, contained at least some minimal references to such tools, being 183 to SacreBLEU and 36 to Moses. Considering that both tools implement the corpus-level aggregation as default, we considered that most of these papers relied on corpus-level BLEU, so there is likely 216 (63\%) papers which implemented corpus-level aggregation for BLEU.

Considering that it is likely that about 2/3 of these papers relied on corpus-level BLEU scores, one first remarkable conclusion is that the results of those papers should be looked with care considering the low statistical significance of corpus-level aggregation. However, we noticed that there are at lest 40 of those papers (about 20\%) that rely on bootstrap resamplings to compute significance tests for system comparisons \cite{koehn-2004-statistical}, which can make the statistical evaluation more robust, even though they in the end reported plain corpus-level BLEU scores \cite{jon-bojar-2023-breeding,fucci-etal-2023-integrating}.




\end{document}